\pdfoutput=1
\documentclass[journal]{IEEEtran}
\usepackage{amsmath,amssymb,amsfonts}
\usepackage{cite}
\usepackage[noend]{algpseudocode}
\usepackage{algorithmicx,algorithm}
\usepackage{algorithm}
\usepackage{graphicx}
\usepackage{lipsum}
\usepackage{enumerate}
\usepackage{amsthm}
\usepackage{color}
\usepackage{booktabs}

\ifCLASSOPTIONcompsoc
  \usepackage[caption=false,font=normalsize,labelfont=sf,textfont=sf]{subfig}
\else
  \usepackage[caption=false,font=footnotesize]{subfig}
\fi

\usepackage{tikz}
\usepackage{textcomp}
\usepackage{hyperref}

\definecolor{BLUE}{HTML}{0000FF}   
\definecolor{ORANGE}{HTML}{FF7F00}   
\definecolor{GREEN}{HTML}{00A877}   
\definecolor{RED}{HTML}{FF0000}   
\definecolor{PURPLE}{HTML}{966FD6}   
\definecolor{BROWN}{HTML}{654321}   
\definecolor{PINK}{HTML}{E7ACCF}   
\definecolor{GREY}{HTML}{7F7F7F}   
\ifCLASSINFOpdf
\else
\fi
\begin{document}

%
\title{Learning Agility and Adaptive Legged Locomotion via Curricular Hindsight Reinforcement Learning}
%
%

\author{
    Sicen Li,
    Yiming Pang, 
    Panju Bai,
    Zhaojin Liu,
    Jiawei Li,

    Shihao Hu,
    Liquan Wang,
    and Gang Wang*
\thanks{This work was supported in part by the National Natural Science Foundation of Heilongjiang Province (Grant No.YQ2020E028). \textit{(Corresponding author: Gang Wang.)}} 
\thanks{
Sicen Li, Yiming Pang, Panju Bai, Shihao Hu, Zhaojin Liu, and Liquan Wang are with College of Mechanical and Electrical Engineering, Harbin Engineering University, Harbin 150001, China: {\tt\small \{ihuhuhu, pangyiming, baipanju, jhjsb, liuzhaojin, wangliquan\}@hrbeu.edu.cn}. 
}
\thanks{Jiawei Li and Gang Wang are with College of Shipbuilding Engineering, Harbin Engineering University, Harbin 150001, China: {\tt\small \{ljw1996, wanggang\}@hrbeu.edu.cn}. }
\thanks{All authors are with the Science and Technology on Underwater Vehicle Laboratory, Harbin Engineering University, Harbin Engineering University, Harbin 150001, China.}

}

\maketitle


\begin{abstract}
Agile and adaptive maneuvers such as fall recovery, high-speed turning, and sprinting in the wild are challenging for legged systems. 
We propose a Curricular Hindsight Reinforcement Learning (CHRL) that learns an end-to-end tracking controller that achieves powerful agility and adaptation for the legged robot. The two key components are (I) a novel automatic curriculum strategy on task difficulty and (ii) a Hindsight Experience Replay strategy adapted to legged locomotion tasks.
We demonstrated successful agile and adaptive locomotion on a real quadruped robot that performed fall recovery autonomously, coherent trotting, sustained outdoor speeds up to 3.45 m/s, and tuning speeds up to 3.2 rad/s.
This system produces adaptive behaviours responding to changing situations and unexpected disturbances on natural terrains like grass and dirt.

\end{abstract}


%
\IEEEpeerreviewmaketitle

%
%
%
%
\section{Introduction}
\IEEEPARstart
Legged systems can execute agile motions by leveraging their ability to reach appropriate and disjoint support contacts, enabling outstanding mobility in complex and unstructured environments\cite{gangapurwala2023learning}. This ability makes them a popular choice for tasks such as inspection, monitoring, search, rescue, or transporting goods in complex, unstructured environments\cite{mitchell2022next}.

However, until now, legged robots could not match the performance of animals in traversing real-world terrains\cite{miki2022learning}. In nature, legged animals such as cheetahs and hunting dogs can make small radius turns at high speed running or recover quickly from a fall with little or no slowing down while chasing prey.
Animals naturally learn to process a wide range of sensory information and respond adaptively in unexpected situations, even when exteroception is completely limited. This ability requires motion controllers capable of recovering from unexpected perturbations, adapting to system and environmental dynamics changes, and executing safe and reliable motion relying solely on proprioception.

Endowing legged robots with this ability is a grand challenge in robotics. Conventional control theories often must be revised to deal with these problems effectively. As a promising alternative, model-free deep reinforcement learning (RL) algorithms autonomously solve complex and high-dimensional legged locomotion problems and do not assume prior knowledge of environmental dynamics. 

Under the privileged learning framework, previous work has solved motion tasks such as fast running\cite{margolis2022rapid}, locomotion over uneven terrain\cite{kumar2021rma}, and mountain climbing\cite{miki2022learning}. However, existing motion controllers can still not respond appropriately and adaptively when various accidents lead to motion interruptions, such as collisions with obstacles or falls caused by stepping into potholes.

In this paper, our goal is to construct a system that can traverse terrains agility and adaptively at an extensive range of commands. Akin to prior work, we found that the policy can be successfully learned when the robot runs with a narrow range of commands and external disturbances. However, when increasing the range of commands or disturbances, the policy propensity to converge to inferior solutions or even diverge as the reward is sparse. The training process can progressively address complex tasks by implementing curriculum learning, but manual curriculum design may also fail due to the difficulty of evaluating learning progress and task difficulty. 
This paper presents Curricular Hindsight Reinforcement Learning (CHRL) to solve this problem by introducing a novel automatic curriculum strategy to automatically assess the learning progress of the policy and control task difficulty. CHRL periodically evaluates the policy's performance and automatically adjusts the curriculum parameters to ramp up task difficulty, including command ranges, reward coefficients, and environment difficulty.

However, adjusting the environment parameters and command ranges destroys the consistency of the state distribution between the environment and the replay buffer, which is disastrous for off-policy RL. This work solved this problem by adapting Hindsight Experience Replay (HER)\cite{andrychowicz2017hindsight} to quadrupedal locomotion tasks to match the existing replay buffer to the current curriculum. Presented HER modifies the commands of past experiences and recalculates rewards, additionally utilizing past experiences, thus increasing the learning efficiency. The proposed policy yields significant performance improvements in learning agility and adaptive high-speed locomotion.

When zero-shot deployed in the real world on uneven outdoor terrain covered with grass, our learned policy sustained a top forward velocity of 3.45 m/s and spinning angular velocity of 3.2 rad/s. The policy also shows strong robustness and adaptability in unstructured environments. When the robot hits an obstacle or steps on an unseen pit and falls, the learned policy shows the capability of failure-resilient running and critical recovery within one second. The robot spontaneously resisted unpredictable external disturbances in indoor tests and demonstrated unique motor skills. These results are reported qualitatively, and corresponding videos highlight the adaptive responses that emerge from end-to-end learning. 

The main contributions of this paper include:
\begin{itemize}
  \item A novel automatic curriculum strategy enables the discovery of behaviors that are challenging to learn directly using reinforcement learning.
  \item We adapt Hindsight Experience Replay to legged locomotion tasks, allowing sample efficient learning and improving performance.
  \item The learned controller can be deployed directly to the real world and performs agility and adaptively in various environments.
\end{itemize}

\section{Related Works}
\label{sec:related_work}
{\bf Model-based Control.}
Dynamic locomotion over unknown and challenging terrain requires careful motion planning and precise motion and force control\cite{humphreys2023bio}.
The primary approach in the legged locomotion community uses model-based mathematical optimization to solve these problems, such as Model-Predictive Control (MPC)\cite{farshidian2017real}, Quadratic Programming\cite{kamidi2022distributed}, and Trajectory Optimisation\cite{buchanan2019walking}. 
The possible contact configurations in legged locomotion greatly expand the search space of trajectories, making the computational cost of optimal planning much higher and making it challenging to meet real-time requirements\cite{ma2022combining}. Modular controller design breaks the control problem into smaller submodules, each based on template dynamics or heuristics. However, these methods are difficult to apply to rough terrain. Combining all models and constraints (whole robot model, contact model, environmental constraints) for optimization through nonlinear MPC\cite{rathod2021model} and whole-body optimization\cite{henze2015approach} can alleviate this problem but is computationally expensive.

{\bf Reinforcement learning for quadrupedal locomotion.}
Instead of tedious manual controller engineering, the Reinforcement learning (RL) technique automatically synthesizes a controller for the desired task by optimizing the controller's objective function\cite{wu2023learning}. 
ANYmal\cite{hwangbo2019learning, hoeller2023anymal} is capable of energy-efficiently following high-level body velocity commands with a max speed of 1.5 m/s, and recovering from falling even in complex configurations.
Extending this approach, \cite{miki2022learning} developed a controller to integrate exteroceptive and proprioceptive perception for legged locomotion, and completed an hour-long hike in the Alps in the time recommended for human hikers.
However, the mechanical design of the ANYmal robot is thought to limit it from running at higher speeds.
\cite{margolis2022rapid} present an end-to-end learned controller that achieves record agility for the MIT Mini Cheetah, sustaining speeds up to 3.9 m/s on flat ground and 3.4 m/s on grassland.
\cite{choi2023learning} demonstrated high-speed locomotion capabilities on deformable terrains. The robot could run on soft beach sand at 3.03 m/s, although the feet were completely buried in the sand during the stance phase.
Compared to previous work, our proposed controller not only realizes high-speed and agile motion on uneven terrains, but also can adaptively handle unexpected events and external disturbances.

{\bf Curriculum learning.}
Prior works have shown that a curriculum on environments can significantly improve learning performance and efficiency using reinforcement learning.
\cite{ji2022concurrent} used a fixed-schedule curriculum on forward linear velocity only.
A Grid Adaptive Curriculum Update Rule is proposed \cite{margolis2022rapid} to track a more extensive range of velocities.
\cite{miki2022learning, rudin2022learning} applied a curriculum on terrains to learn highly robust walking controllers on non-flat ground.
\cite{nahrendra2023dreamwaq} utilized a game-inspired curriculum to ensure progressive locomotion policy learning over rugged terrains.
Compared to previous work, our approach adjusts command, reward coefficients, and environment difficulty in stages, allowing for finer control of curriculum difficulty. The combination of curriculum learning and HER also further enhances learning efficiency.

{\bf Hindsight Experience Replay.}
Hindsight Experience Replay (HER) has paved a promising path toward increasing the efficiency of goal-conditioned tasks with sparse rewards\cite{andrychowicz2017hindsight}. Several variants have been proposed to enhance HER, such as curriculum learning to maximize the diversity of the achieved goals\cite{fang2019curriculum}, providing demonstrations\cite{8463162}, generating more valuable desired goals\cite{han2023overfitting}, and curiosity-driven exploration\cite{9197421}.
Legged locomotion tasks can be viewed as a particular variant of goal-conditioned tasks due to the presence of control commands. This work adapts HER to legged locomotion tasks and proves its effectiveness.

\section{Method}
\label{sec:Method}
Our goal is to learn a policy that takes sensory data and velocity commands as input and gives as output joint desired positions. The command $\mathbf{c}^{cmd}_t$ includes the linear velocity $\mathbf{v}^{cmd}_t$ and its yaw direction $\mathbf{d}^{cmd}_t$. 

The policy is trained in simulation and then performs zero-shot sim-to-real deployment. 
The controller is trained via privileged learning, which consists of two stages:

First, a teacher policy $\pi_{\theta}^{teacher}$ with parameters $\theta$ is trained via reinforcement learning with full access to privileged information that includes the ground-truth state of the environment. Proposed method Curricular Hindsight Reinforcement Learning
(CHRL) works in this stage and improves the performance of the teacher policy. Second, a student policy $\pi_{\phi}^{student}$ is trained via imitation learning to predict the teacher’s optimal action given only partial and noisy observations of the environment. Then, the student policy is deployed on the robot without any fine-tuning.

\subsection{Training environment}
We use pybullet\cite{coumans2021} as our simulator to build the training environment. 
A procedural terrain generation system is developed to generate diverse sets of trajectories. Four parallel agents collect five million simulated time steps for policy training, which spends less than 6 hours of wall-clock time using a single NVIDIA RTX 3090 GPU.

\subsection{Hardware}
The robot used in this work stands 35 cm tall and weighs 15 kg. The robot boasts 18 degrees of freedom (DoFs) with 12 actuated joints, each capable of delivering a maximum torque of 33.5 Nm and six generalized coordinates for the floating base. 
Our neural network controller runs at 66 Hz on an onboard NVIDIA Jetson AGX Orin computer. 
The robot is equipped with an Inertial Measurement Unit (IMU) and joint encoders.

\subsection{Control Architecture}
\textbf{Action space.} The action $\mathbf{a}_t$ is a 12-dimensional desired joint position vector. A PD controller is used to calculate torque $\boldsymbol{\tau} = Kp(\hat{\mathbf{q}} - \mathbf{q})+Kd(\hat{\dot{\mathbf{q}}} - \dot{\mathbf{q}})$, $Kp$ and $Kd$ are manually specified gains, which are set to 27.5 and 0.5 respectively. The target joint velocities $\hat{\dot{\mathbf{q}}}$ are set to 0.

\textbf{Student Observation Space.} The robot’s joint encoders provide joint angles $\mathbf{q}_{t} \in \mathbb{R}^{12} $ and velocities $\dot{\mathbf{q}}_{t} \in \mathbb{R}^{12}$. $\mathbf{g}_{t}^{\text {ori}} \in \mathbb{R}^{3}$ and $\omega_{t}^{\text {ori}} \in \mathbb{R}^{3}$ denote the orientation and angular velocities measured using the IMU. $\pi_{\phi}^{student}$ takes as input a history of previous observations and actions denoted by $\mathbf{o}_{t-H:t}$ where
$\mathbf{o}_{t}=\left[\mathbf{q}_{t}, \dot{\mathbf{q}}_{t}, \mathbf{g}_{t}^{\text {ori}}, \omega_{t}^{\text {ori}}, \mathbf{a}_{t-1}\right]$. The input to $\pi_{\phi}^{student}$ is $\mathbf{x}_{t}= (\mathbf{o}_{t-H:t}, \mathbf{c}_{t})$, where $\mathbf{c}_{t}$ is specified by a human operator through remote control during deployment, $H=100$. 

\textbf{Teacher Observation Space.} The teacher observation is defined as $\mathbf{s}_{t} = (\mathbf{o}_{t-H:t}, \mathbf{c}_{t}, \mathbf{p}_{t-H:t})$, where $H=4$. $\mathbf{p}_t$ is the privileged state which contains the body velocity $\mathbf{v}_{t} \in \mathbb{R}^{3}$, the binary foot contact indicator vector $\mathbf{f}^{t} \in \mathbb{R}^{4}$, the relative position in the world frame $\mathbf{p}_{t} \in \mathbb{R}^{3}$, friction coefficient of feet, and payload mass.

\textbf{Reward Function.}\label{sec:Reward} The reward function encourages the agent to move forward and penalizes it for jerky and inefficient motions. 
We denote joint torques as $\boldsymbol{\tau}_{t} \in \mathbb{R}^{12}$, the acceleration of the base in the base frame of the robot as $\ddot{a}_{t} \in \mathbb{R}^{3}$, the velocity of the feet as $\mathbf{v}_{t}^{f} \in \mathbb{R}^{4}$, and the total mechanical power of the robot as $\mathbb{W}_{t}$.
The details of the reward function:
\begin{itemize}
    \item Linear Velocity: $\exp \left\{-0.5\left(\mathbf{v}_{t}^{cmd}-\mathbf{v}^{x}_{t}\right)^{2}\right\}$
    \item Linear Velocity Penalties: $ - 0.4\mathbf{v}^{y}_{t} - 0.4\mathbf{v}^{z}_{t}$
    \item Orientation Tracking: $\exp \left\{-0.5\left(\mathbf{d}_{t}^{cmd}-\mathbf{g}_{t}^{\mathrm{ori}}\right)^{2}\right\}$
    \item Height Constraint: $-\lvert \mathbf{p}^{z}_{t}-\mathbf{p}^{\mathrm{target}}_{t} \rvert$
    \item Angle Velocity Penalties: $-\left\| \omega_{t}^{\text {ori}} \right\|^{2}$
    \item Self-collision Penalties: $-\mathbf{1}_{\mathrm{selfcollision}}$
    \item Joint Torque Penalties: $-\left\| \boldsymbol{\tau}_{t} \right\|^{2}$
    \item Base Acceleration Penalties: $-\left\| \ddot{a}_{t} \right\|^{2}$
    \item Energy: $-\mathbb{W}_{t}$
    \item Foot Slip: $-\left\|\operatorname{diag}\left(\mathbf{f}^{t}\right) \cdot \mathbf{v}_{t}^{f}\right\|^{2}$
\end{itemize}

The reward at time $t$ is defined as the sum of the quantities with the scaling factor of each term being 3.0, 1.0, 3.0, 10, 0.21, 2.0, 0.018, 0.1, 0.012, and 0.3.

\subsection{Policy Optimization}

\textbf{Teacher Policy.} The teacher policy $\pi_{\theta}^{teacher}$ is modeled as an MLP, which consists of two MLP components: a state encoder $g_{\theta_{e}}$ and the main network $\pi_{\theta_{m}}$, such that $\mathbf{a}_{t}=\pi_{\theta_{m}}\left(\mathbf{z}_{t}\right)$ where $\mathbf{z}_{t}=g_{\theta_{e}}\left(\mathbf{s}_{t}\right)$ is a latent representation. Each module is parameterized as a neural network with $256$ hidden nodes respectively and rectified linear units (ReLU)\cite{nair2010rectified} between each layer. We optimize the teacher parameters together using REDQ\cite{chen2020randomized}.

\textbf{Student Policy.} We use the same training environment as for the teacher policy, but add additional noise to the student observation $\mathbf{o}_{t}^{noise} = n(\mathbf{o}_{t})$ where $n(\mathbf{o}_{t})$ is a Gaussian noise model applied to the observation. The student policy uses a temporal convolutional network (TCN) \cite{lee2020learning} encoder $g_{\phi_{e}}$ to solve the Partially Observable Markov Decision Process (POMDP). The student action $\hat{\mathbf{a}}_{t} = \pi_{\phi_{m}}(\hat{\mathbf{z}}_{t})$ where $\hat{\mathbf{z}}_{t}=g_{\phi_{e}}\left(\mathbf{o}_{t}^{noise}\right)$. The student policy is trained via supervised learning. The loss function is defined as 
\begin{equation}
\label{eq:loss}
\begin{aligned}
  \mathcal{L}=(\hat{\mathbf{a}}_{t} - \mathbf{a}_{t})^{2} + (\hat{\mathbf{z}}_{t} - \mathbf{z}_{t})^{2}.
\end{aligned}
\end{equation}
We employ the dataset aggregation strategy (DAgger)\cite{ross2011reduction}. Training data are generated by rolling out trajectories according to the student policy. The weights of all networks are initialized with Kaiming Uniform Initialization\cite{he2015delving}, and the biases are zero-initialized. All parameters are updated by the Adam optimizer~\cite {kingma2014adam} with a fixed learning rate $3 \times 10^{-4}$.

\subsection{Domain Randomization}
Domain randomization encourages the policy to learn a single behavior that works across all the randomized parameters to cross the sim-to-real gap. We apply external force and torque to the robot's body, introduce slippage by setting the friction coefficients of the feet to a low value, and randomize the robot's dynamics parameters\cite{tan2018sim, miki2022learning}. Before each training episode, we randomly select a set of physical parameters (Table \ref{table:random parameter}) to initiate the simulation. 
\begin{table}
\caption{Ranges of the randomized parameters.}
\label{table:random parameter}
\centering
\small
\begin{tabular}{l|ccc}
\toprule
Term & Min & Max & Unit\\
\midrule
Contact friction & 0.4 & 1.25 & - \\
Payload mass & -1.0 & 5.0 & kg \\
Robot mass bias & 85\% & 115\% & -\\
Motor friction & 0 & 0.05 & Nm \\
Control frequency & 60 & 70 & Hz \\
Latency & 0 & 40 & ms \\
IMU bias & -0.05 & 0.05 & rad \\
IMU noise (std) & 0 & 0.05  & rad \\
Motor encoder bias & -0.08 & 0.08 & rad \\
External disturbance force & 0 & 15 & N \\
External disturbance torque & 0 & 5 & Nm \\
Gravity direction shift & 0 & 15 & degree \\
\toprule
\end{tabular}
\end{table}

\subsection{Automatic Curriculum Strategy}\label{sec:curriculum}
Similar to previous work, our approach implements a training curriculum that progressively modifies the distribution of environmental parameters, thus enabling policies to improve motor skills and continuously adapt to new environments.

Some works use a curriculum where the commands are updated on a fixed schedule $\mathbb{C}^{k}$, as a function of the timing variable $k$. The schedule $\mathbb{C}^{k}$ consists of two parts, the distribution $p_{x}^{k}$ of the random variable $x$  (such as domain randomization parameters and commands) and $r_{c}^{k}$ for the curriculum coefficient $c$ (such as reward factors).

The update rule $f$ takes the form $\mathbb{C}^{k+1} \longleftarrow f(\mathbb{C}^{k})$. However, a fixed schedule requires manual tuning. If the environment, rewards, or learning settings are modified, the schedule will likely need to be re-turned, which will be costly in terms of time. Rather than advancing the curriculum on a fixed schedule, we automatically update the curriculum using a command-based rule.


Unlike previous works, our approach is not limited to the use of commands or terrain but can take advantage of more environmental parameters, such as reward coefficients and domain randomization parameters, to control the task's difficulty at a finer granularity. 
Table \ref{table: curriculum parameter} shows the parameters used in our experiences. 

\begin{table}
\caption{Range of curriculum parameters.}
\label{table: curriculum parameter}
\centering
\begin{tabular}{l l|ccc}
    \toprule
    \multicolumn{2}{l|}{} & Start & End & Unit\\
    \midrule
    \multicolumn{2}{l|}{\it{Reward factor}}& \\
        & Angle Velocity & 0 & 0.21 & -\\
        & Joint Torque Penalties & 0 & 0.018 & -\\
        & Base Acceleration Penalties & 0 & 0.1 & -\\
        & Energy & 0 & 0.012 & -\\
        & Foot Slip & 0 & 0.3 & -\\
    \midrule
        \multicolumn{2}{l|}{\it{Domain Randomization}}& \\
        & Contact friction & $[1.25, 1.25]$ & $[0.4, 1.25]$ & -\\
        & Payload mass & $[-1.0, -1.0]$ & $[-1.0, 5.0]$ & kg\\
        & External disturbance force & $[0, 0]$ & $[0, 15]$  & N\\
        & External disturbance torque & $[0, 0]$ & $[0, 5]$  & Nm\\
        & Gravity direction shift & $[0, 0]$ & $[0, 15]$ & degree\\
    \midrule
        \multicolumn{2}{l|}{\it{Terrians}}& \\
        & Height fields range & $[0, 0]$& $[-2.5, 2.5]$ & cm\\
    \midrule
        \multicolumn{2}{l|}{\it{Commands}}& \\
        & linear velocities & $[0, 0.8]$& $[0, 3.5]$ & m/s\\
        & Direction of velocities & $[-0.5, 0.5]$& $[-\pi, \pi]$ & rad\\
    \toprule
    \end{tabular}
\end{table}

In this work, we apply a tabular Curriculum Update Rule. First, we manually set the number of curriculums $N$ to uniformly split the curriculum parameters (Table \ref{table: curriculum parameter}) from start to end into $p_{x}^{1} \dots p_{x}^{N}$ and $r_{c}^{1} \dots r_{c}^{N}$.
At episode $k$, the curriculum parameters for the agent and environment are sampled from the joint distribution $p_{x}^{n}$, and the reward factor is replaced with $r_{c}^{n}$. If the agent succeeds in this region of the curriculum space, we would like to increase the difficulty by updating the tabular curriculum from $p_{x}^{n}$ and $r_{c}^{n}$ to $p_{x}^{n+1}$ and $r_{c}^{n+1}$:

\begin{equation}
\label{eq: evaluation metric}
\centering
\begin{aligned}
    p_{x}^{n}, r_{c}^{n} \leftarrow\left\{\begin{array}{ll} p_{x}^{n+1}, r_{c}^{n+1} & \epsilon_{k}\left[\mathbf{v}^{\mathrm{cmd}}\right] < \epsilon_{0} \text{ and } \epsilon_{k}\left[\mathbf{d}^{\mathrm{cmd}}\right] < \epsilon_{1} \\
p_{x}^{n}, r_{c}^{n} & \text { otherwise }
\end{array}\right.
\end{aligned}
\end{equation}

where 
\begin{equation}
\label{eq: evaluation metric2}
\centering
\begin{aligned}
    \epsilon_{k}\left[\mathbf{v}^{\mathrm{cmd}}\right]=\mathbb{E}_{\mathbf{v}_{t}^{\mathrm{cmd}} , \boldsymbol{d}_{t}^{\mathrm{cmd}}} \sqrt{\mathbb{E}_{t}\left(\mathbf{v}_{t}^{\mathrm{cmd}}-\mathbf{v}_{t}^{x}\right)^{2}},
  \\
    \epsilon_{k}\left[\mathbf{d}^{\mathrm{cmd}}\right]=\mathbb{E}_{\mathbf{v}_{t}^{\mathrm{cmd}} , \boldsymbol{d}_{t}^{\mathrm{cmd}}} \sqrt{\mathbb{E}_{t}\left(\mathbf{d}_{t}^{\mathrm{cmd}}-\mathbf{g}_{t}^{\mathrm{ori}}\right)^{2}},
\end{aligned}
\end{equation}
$\mathbf{v}_{t}^{\mathrm{cmd}}$ and $\mathbf{d}_{t}^{\mathrm{cmd}}$ are the linear velocity of the robot and its yaw direction measured at time $t$. 
The average tracking error $\epsilon_{k}\left[\mathbf{v}^{\mathrm{cmd}}\right]$ and $\epsilon_{k}\left[\mathbf{d}^{\mathrm{cmd}}\right]$ over trials with the current policy is defined as the main evaluation metric.

Measuring both of these metrics individually does not effectively reflect the performance of the controller, so we need a performance metric that measures both commands. We construct a composite metric that captures the diversity of commands that the controller can execute within a certain maximum error tolerance:
\begin{equation}
\label{eq: evaluation metric3}
\centering
\begin{aligned}
    \epsilon_{k}\left[\mathbf{v}^{\mathrm{cmd}}\right] < \epsilon_{0} \text{ and } \epsilon_{k}\left[\mathbf{d}^{\mathrm{cmd}}\right] < \epsilon_{1}
\end{aligned}
\end{equation} 

The proposed automatic Curriculum strategy cyclical evaluates the learned policy and automatically controls curriculum difficulty based on evaluation results. The evaluation procedure will be initiated for every 100,000 samples collected by the agent. Ten evaluation processes were run in parallel, each lasting 10 seconds and randomly sampling new commands every 2 seconds. In all experiences, we set $\epsilon_{0} = 0.15$, $\epsilon_{1} = 0.25$, and the number of curriculums $N=10$.

\begin{figure*}[ht]
\centering
    \includegraphics[width=0.99\textwidth]{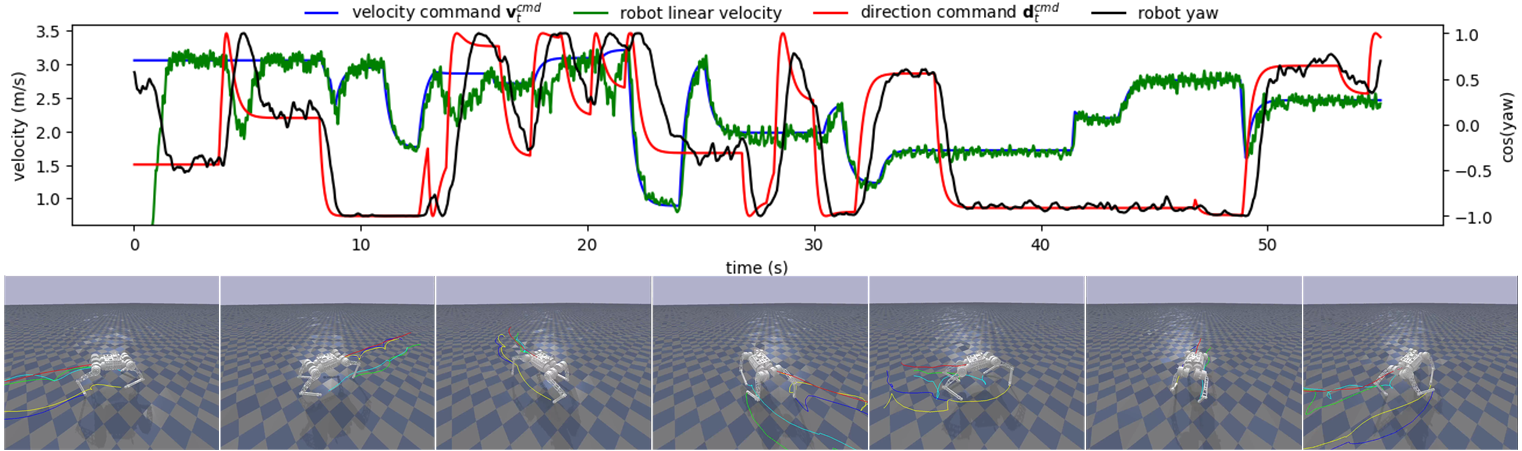}
    \caption{Command tracking test results in simulation.}
    \label{fig-simulation}
\end{figure*}

\subsection{Hindsight Experience Replay for Legged Locomotion}
\begin{algorithm}[t]
\caption{Curricular Hindsight Reinforcement Learning} \label{alg:chrl}
\begin{algorithmic}[1]
\State Initialize an off-policy RL algorithm $\mathbb{A}$, a reward function $r$, replay buffer $\mathcal{B}$, an curriculm schedule $\mathbb{C}$, training environment $\mathbb{E}$.

\For{each iteration}
    \If {$\mathbb{C}$ need update}
        \State Evaluate and get $\epsilon_{k}\left[\mathbf{v}^{\mathrm{cmd}}\right] \text{ and } \epsilon_{k}\left[\mathbf{d}^{\mathrm{cmd}}\right]$ \eqref{eq: evaluation metric2}
        \State Update $\mathbb{C}$ according to rule \eqref{eq: evaluation metric}
        \State Update parameters of environment $\mathbb{E}$ with $\mathbb{C}$
    \EndIf
        
    \State Get a state $\mathbf{s}_t$ from $\mathbb{E}$ and sample a command $\mathbf{c}_{t}^{cmd}$
    \State Execute an action $\mathbf{a}_t \sim \pi_{\theta}^{teacher}\left(\cdot \mid \mathbf{s}_t, \mathbf{c}_{t}^{cmd} \right)$
    \State Observe reward $r_t$, new state $\mathbf{s}_{t+1}$ and robot state $\mathbf{s}^{robot}_{t+1}$ \eqref{eq:transition}
    \State \textsc{// Storing phase}
    \State Store transition tuple:
    \State $\mathcal{B} \leftarrow \mathcal{B} \cup\left(\mathbf{s}_t, \mathbf{a}_t, r_{t}, \mathbf{s}_{t+1}, \mathbf{c}_{t}^{cmd}, \mathbf{s}^{robot}_{t+1}\right)$

    \State \textsc{// training phase}
    \State Sample random minibatch $B$ from $\mathcal{B}$:
    \State $\{\mathbf{s}_j, \mathbf{a}_j, r_{j}, \mathbf{s}_{j+1}, \mathbf{c}_{j}^{cmd}, \mathbf{s}^{robot}_{j+1}\}_{j=1}^B$

    \State Resample command $\mathbf{c}_{j}^{cmd} \longleftarrow \mathbf{c}_{new}^{cmd}$\eqref{eq:command}
    \State Recalculate $r_{j}$ with $c_{j}^{cmd}$ and $\mathbb{C}$
    \State Perform one step of optimization using $\mathbb{A}$ and minibatch $B$
\EndFor 

\end{algorithmic}
\end{algorithm}

Hindsight Experience Replay (HER)\cite{andrychowicz2017hindsight} allows the algorithm to reuse existing samples and can be combined with any off-policy RL algorithm. However, the original HER algorithm is only suitable for some goal-conditioned tasks with sparse and binary rewards. We have made some improvements to HER to accommodate legged locomotion tasks. 

Firstly, we sample a new goal for each transition during the training phase instead of choosing a different goal for a trajectory during the storing phase. 
Second, due to the need to recalculate rewards, more information about the state of the robot ($\mathbb{W}, \ddot{a}, \omega^{\text {ori}}, \tau, \mathbf{f}, \mathbf{v}, \mathbf{g}^{\mathrm{ori}}$) is added in each transition. Then the transition $\mathbf{T}_t$ transition becomes
\begin{equation}
\label{eq:transition}
\centering
\begin{aligned}
    &\mathbf{T}_t = \left(\mathbf{s}_t, \mathbf{a}_t, r_{t}, \mathbf{s}_{t+1}, \mathbf{c}_{t}^{cmd}, \mathbf{s}^{robot}_{t+1} \right), \\
    &\mathbf{s}^{robot}_{t+1} \doteq \{\mathbb{W}_{t+1}, \ddot{a}_{t+1}, \omega_{t+1}^{\text {ori}}, \tau_{t+1}, \mathbf{f}^{t+1}, \mathbf{v}_{t+1}, \mathbf{g}_{t+1}^{\mathrm{ori}}\}
\end{aligned}
\end{equation} 

Third, the original HER samples the goal achieved in the episode's final state. This is inefficient as policies for similar commands in locomotion tasks tend to be similar. For a transition with command $\mathbf{c}_{t}^{cmd}$ consisted with the linear velocity $\mathbf{v}^{cmd}_t$ and its yaw direction $\mathbf{d}^{cmd}_t$. We sample new command $\mathbf{c}_{new}^{cmd}$ by adding neighboring regions to the sampling distribution:
\begin{equation}
\label{eq:command}
\centering
\begin{aligned}
    & \mathbf{c}_{new}^{cmd} \doteq \{\mathbf{v}^{cmd}_{new}, \mathbf{d}^{cmd}_{new}\}, \\
    & \mathbf{v}^{cmd}_{new} \sim \mathcal{U}[\mathbf{v}^{cmd}_t - 0.3, \mathbf{v}^{cmd}_t + 0.3], \\
    & \mathbf{d}^{cmd}_{new} \sim \mathcal{U}[\mathbf{d}^{cmd}_t - 0.5, \mathbf{d}^{cmd}_t + 0.5].
\end{aligned}
\end{equation} 
All new commands sampled will be limited to the range of the current curriculum. Then $r_t$ is recalculated with $\mathbf{s}^{robot}_{t+1}$ according to Section \ref{sec:Reward}.

This approach may be seen as a form of data augmentation as samples generated by the policy of a specific command are shared within neighboring policies. Combined with the automatic curriculum strategy in Section~\ref{sec:curriculum}, we present Curricular Hindsight Reinforcement Learning (CHRL). The pseudocode for CHRL is shown in Algorithm \ref{alg:chrl}.

\section{Results}

\subsection{Command Tracking Test}
\begin{figure}[ht]
\centering
    \includegraphics[width=0.49\textwidth]{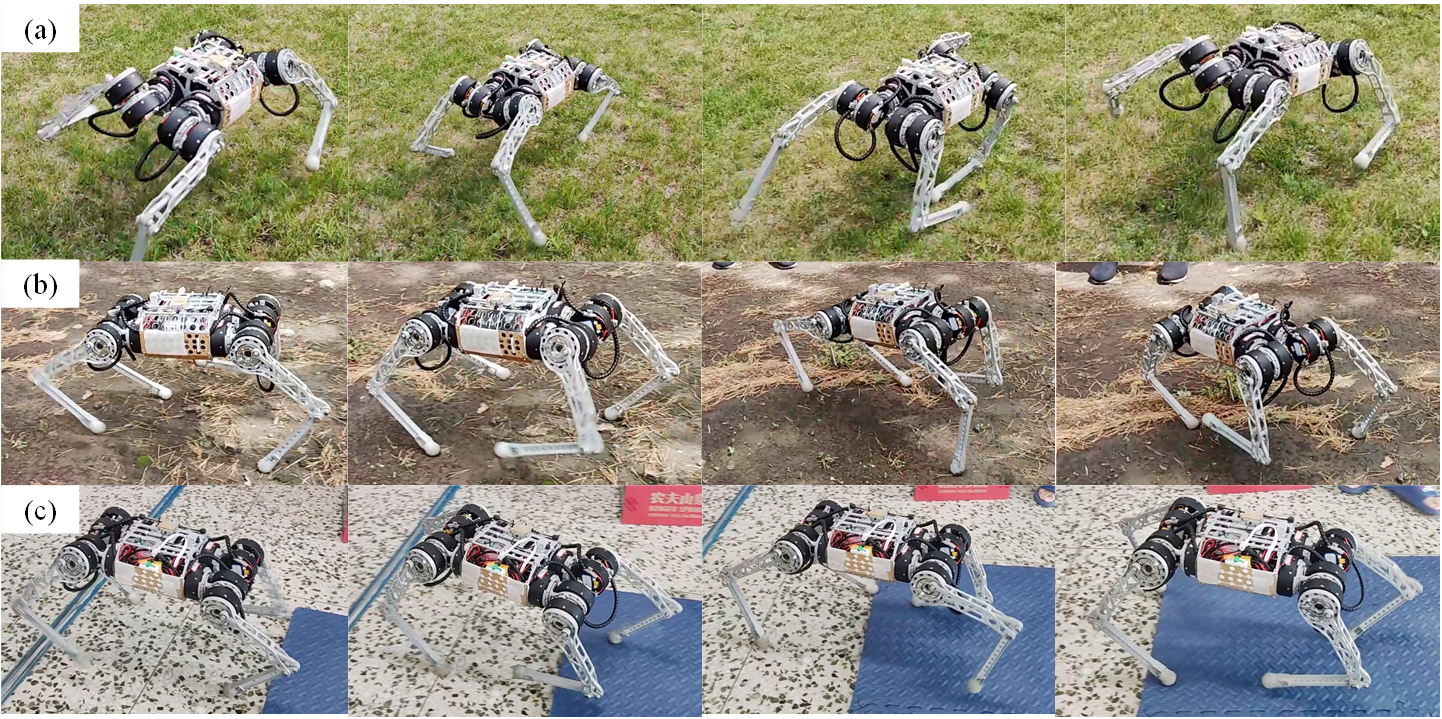}
     \caption{Real Robot Deployment Experiments. (a) Running on grassland. (b) Walking on wet dirt ground. (c) Indoor walking through different ground conditions.}
    \label{fig-terrians}
\end{figure}

Firstly, the control performance was evaluated in forward running under random commands in simulation. In our experiments, we resampled commands and sent them to the robot with a probability of 1/150 (every 2.25 seconds on average) and resampled environmental variables with the same probability.
Fig. \ref{fig-simulation} shows the command tracking accuracy of the policy in the simulation. 

The robot can move steadily and well in the desired direction in rough terrain. Even if the velocity command changes during direction correction, the controller can track both commands well. Note that the observed velocity oscillation around the commanded velocity is a well-known phenomenon in legged systems, including humans \cite{winter1991biomechanics}. Regarding the average speed, the learned policy has an error of less than 5\% on the simulated robot.

We also performed experiences with real robots, including speed tracking tests at 0.8 m/s indoors and a maximum of 3.5 m/s outdoors. The outdoor terrain presented multiple challenges not present in indoor running, including variations in ground height, friction, and terrain deformation. With these variations, the robot must actuate its joints differently to achieve high velocities that are different from those achieved on flat, rigid terrain with high friction, such as a treadmill or paved road. We estimated the robot's locomotion speed by measuring the time it took to pass through a 5-meter-long section of the road and performed multiple sets of repetitive tests. In the indoor 0.8 m/s speed test, we recorded a 5 m walking time of 6.17 s, with an average speed of 0.81 m/s. In contrast, we recorded an outdoor 5-meter sprint time of 1.45 seconds with an average speed of 3.45 m/s. The results show that the robot can track speed commands consistently and accurately in unknown scenarios, both indoors and outdoors.

As shown in Fig. \ref{fig-terrians}, we evaluated the yaw tracking control of the controller in an outdoor grass environment. In the test, the robot's desired direction command and velocity command were randomly sent by the human operator. Our experiments observed a maximum yaw speed of 3.2 rad/s for the robot, followed by a safe stop. Even with continuous slewing motion at larger velocity commands (3 m/s), the robot remained stable while turning, demonstrating the submissive interactions and robustness learned by the agent.

The robot could track commands robustly in different ground conditions with different hardness, friction, and obstacles (Fig. \ref{fig-terrians}). The learned motor skills were stable across the different ground conditions, and the robot continued to trot steadily in all three conditions. The trained policy exhibited compliant interaction behaviors to handle physical interactions and impacts.
In testing, when a large foot slip occurred, the robot could recover quickly, even running near maximum speed. If the command was suddenly set to zero during operation, the robot assumed a stable posture and quickly stopped moving.








\subsection{Fall Recovery and Response to Unexpected Disturbances}

\begin{figure}[ht]
\centering
    \includegraphics[width=0.485\textwidth]{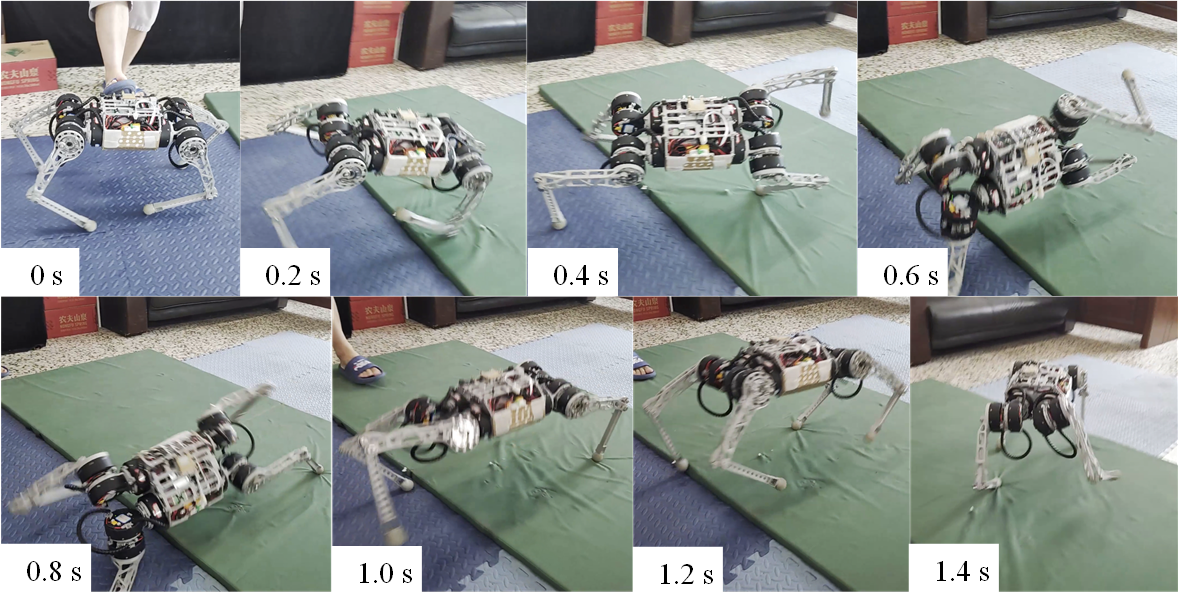}
     \caption{Indoor walking with unexpected disturbance. The robot was suddenly kicked when it was walking indoors, causing it to lose its balance. The controller controlled the legs to support the body and restore balance in one second.}
    \label{fig-external}
\end{figure}

\begin{figure}[ht]
\centering
    \includegraphics[width=0.485\textwidth]{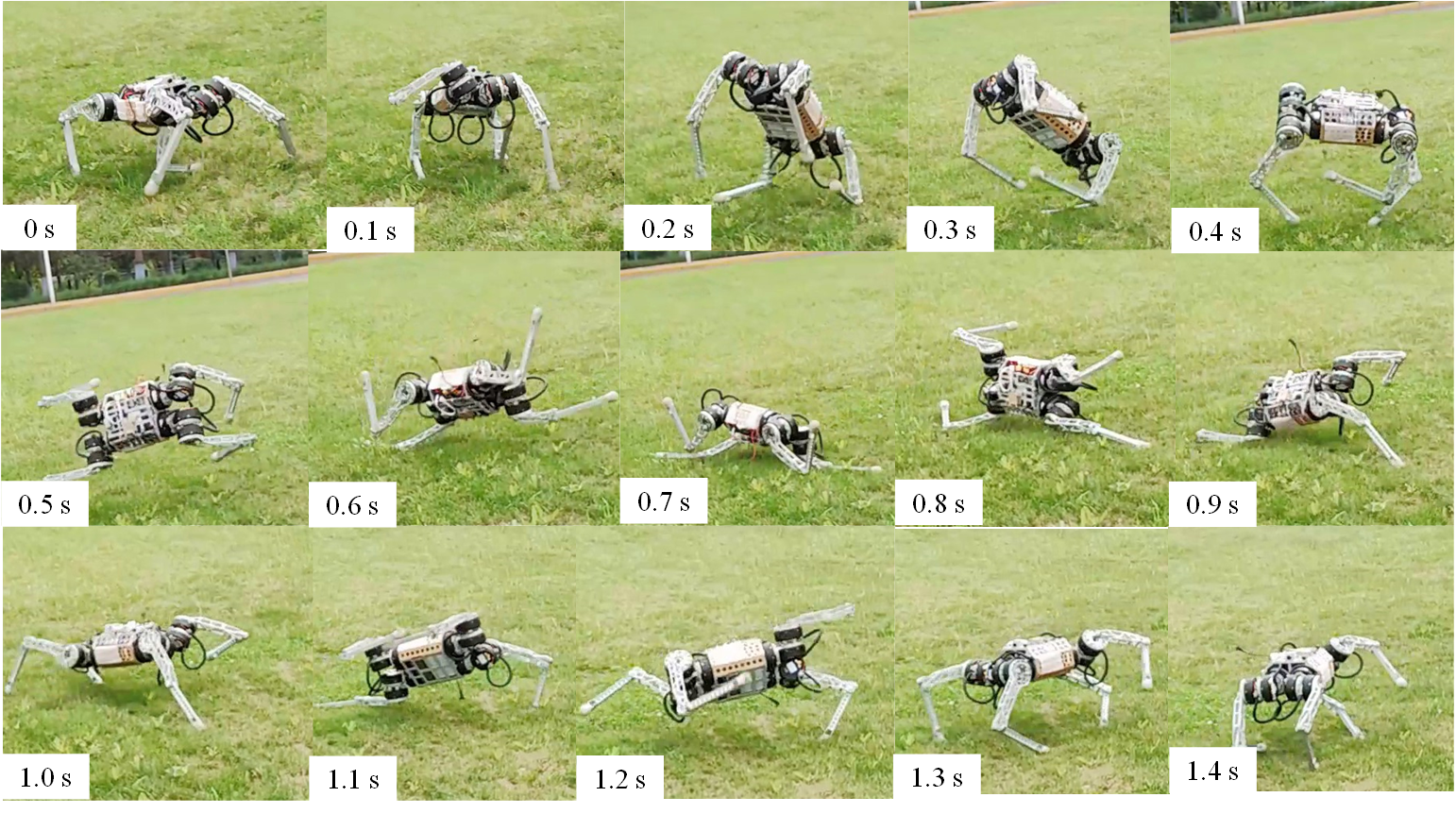}
     \caption{Outdoor running accident. The robot ran on grassland with command 3m/s when it stepped on an unknown deep pit, causing it to trip and fall. The controller utilized forward inertia to roll the body and control the legs in an impact-resistant stance to protect the robot, followed by a quick return to running.}
    \label{fig-grassfall}
\end{figure}

\begin{figure}[ht]
\centering
    \includegraphics[width=0.485\textwidth]{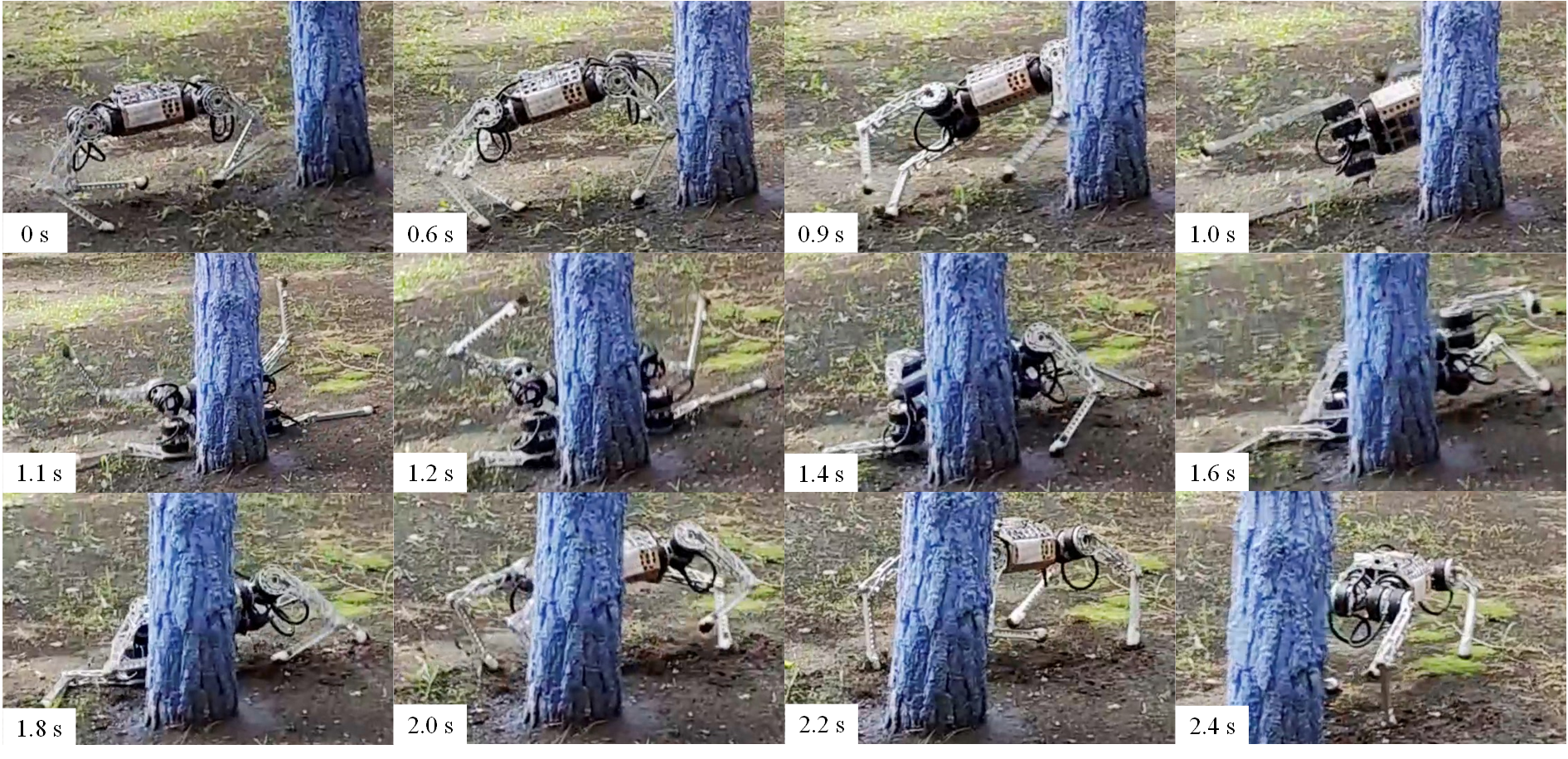}
     \caption{Outdoor fall recovery test. In the test, we deliberately controlled the robot to hit a tree on a dirt floor, and due to the tree's obstruction, the robot could not move forward and fell. The robot actively contacted the ground with its legs to support its body and held it upright.}
    \label{fig-fall2}
\end{figure}

Due to the uncertainties in unforeseen situations, locomotion failures are likely to occur. We illustrate these challenges in robotic locomotion using field tests (Fig. \ref{fig-grassfall} and \ref{fig-fall2}) and adaptive behaviors that are robust to uncertainty (Fig. \ref{fig-external}).

During the outdoor tests, we found that the robot also experienced many unexpected accidents, which resulted in unintended contact of the robot's body with the environment.
In the indoor tests, we actively applied an external disturbance to the robot during its walking to destabilize its movement and observe its reaction after losing balance. Typically, robots fall within a second of losing their balance, and the window of time to prevent a fall is about 0.2 - 0.5 seconds. The proposed controller is observed to have different adaptive behaviors for the above unexpected scenarios, which generate dynamic locomotions and complex leg coordination for immediate recovery from failures.

In these unexpected scenarios, our robots autonomously coordinate different locomotion patterns to mitigate disturbances and prevent or recover from failures without human assistance. These behaviors are extremely similar to the behavior of biological systems (such as cats, dogs, and humans), which shows higher versatility and intelligence: the ability to deal with constantly changing and complex situations.

We classified learned response behaviors into three strategies:
\begin{itemize}
  \item Natural rolling using semi-passive motion (Fig. \ref{fig-grassfall}). Natural rolling is the behavior of a robot that uses its inertia and gravity to tumble.
  \item Active righting and tumbling (Fig. \ref{fig-fall2}). Active righting is a policy in which the robot actively uses its legs and elbows to propel itself and generate momentum to flip into a prone position.
  \item Stepping. Fig. \ref{fig-external} shows an example of stepping, where coordination and switching of the support legs are required to regain balance in the event of a loss of stability of the current motion due to an external disturbance. This multi-touch switching occurs naturally using a learning-based policy.
\end{itemize}

Compared to a manually designed fall recovery controller with a fixed pattern, our learned controller can recover from various fall situations by responding to dynamic changes using online feedback. In contrast, a manual controller can only cope with a narrow range of situations.









\subsection{Analysis of Skill Adaptation}

\begin{figure}[ht]
\centering
    \includegraphics[width=0.45\textwidth]{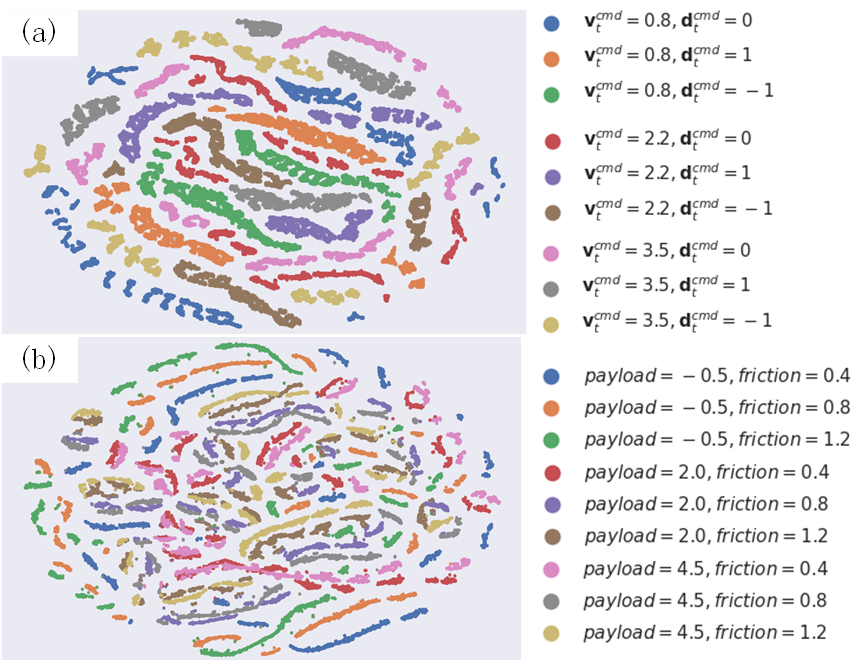}
     \caption{Two-dimensional t-SNE embedding of the representations in the last hidden layer of the student networks. (a) Embedding under different commands. (b) Embedding under different robot payloads (kg) and ground friction.}
    \label{fig-t-SNE}
\end{figure}

We analyzed the features learned by the policies separately using t-distributed Stochastic Neighbour Embedding (t-SNE), thus investigating how skills are adapted and distributed in the network. The t-SNE algorithm is a dimensionality reduction technique used to embed high-dimensional data into a low-dimensional space and to visualize it. Similar features in the output of the student network will appear with high probability in the same neighborhood of the clustered points (Fig. \ref{fig-t-SNE}) and vice versa.

A two-dimensional projection of the student network features by t-SNE visualizes the neighborhoods and clusters of the samples. In Fig. \ref{fig-t-SNE}(a), t-SNE analyses of student network features under different commands show that the agent learns unique skills and patterns for different commands, revealing a diversity of skills after course training. If the commands are very similar, the network reuses certain patterns and features to some extent but also fine-tunes for these minor differences.

We also use t-SNE to compare policy features with different payloads and friction coefficients. As shown in Fig. \ref{fig-t-SNE}(b), the features under different environmental variables are far away from each other, meaning that the student policy can clearly recognize subtle environmental changes and respond appropriately.




\subsection{Comparative Evaluation}
\begin{figure}[ht]
\centering
    \includegraphics[width=0.485\textwidth]{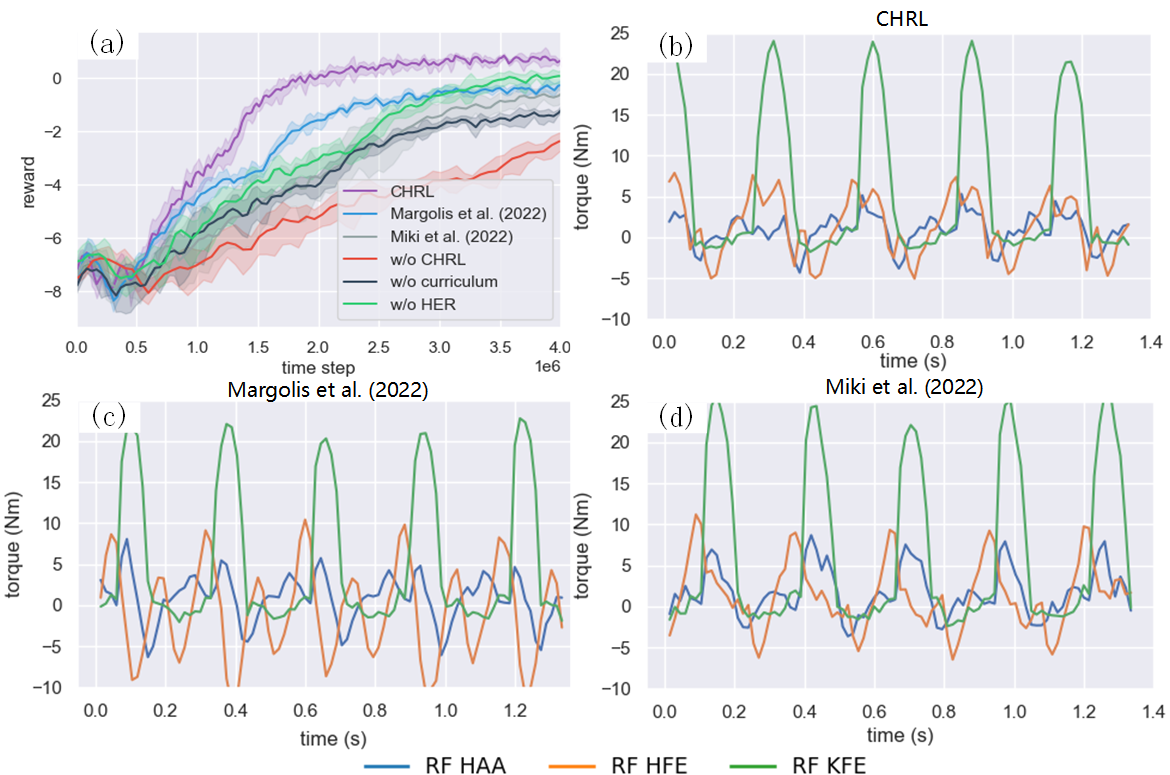}
     \caption{(a) Learning curves for different curriculum strategies. The horizontal axis indicates the number of time steps. The vertical axis shows the average reward over time steps within an episode. The shaded areas denote one standard deviation over four runs. (c-d) The measured torques of the right leg during forward running (around 3.0 m/s) for different curriculum strategies.}
    \label{fig-Comparison}
\end{figure}

\begin{figure}[ht]
\centering
    \includegraphics[width=0.485\textwidth]{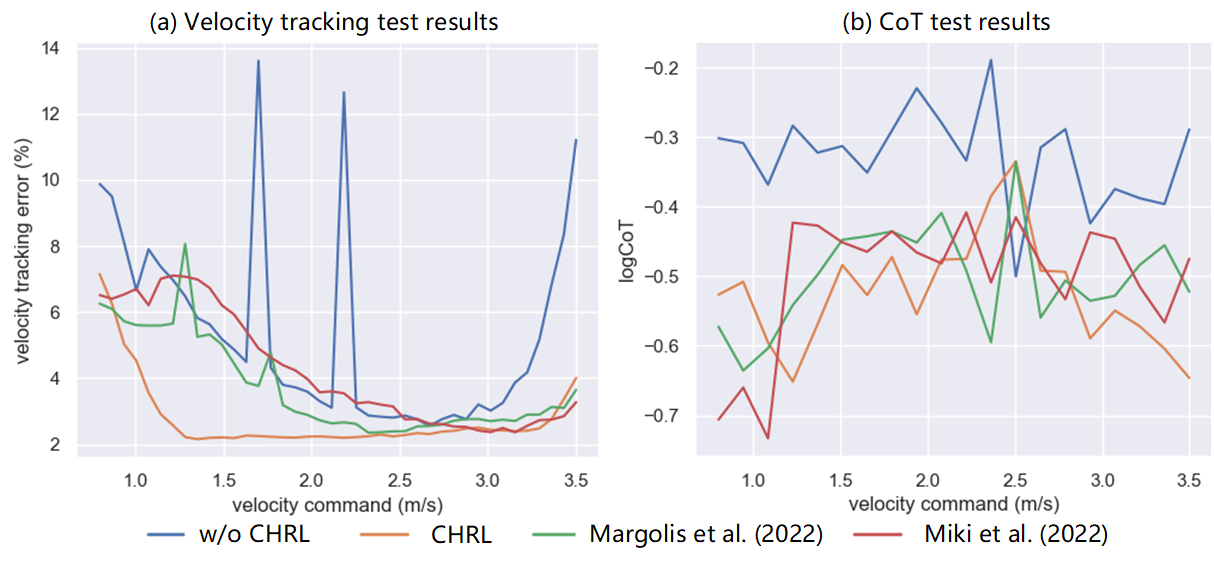}
     \caption{Velocity tracking test and CoT test results averaged over four runs.}
    \label{fig-ablation}
\end{figure}
We compare the performance with the following baselines:
\begin{itemize}
  \item The grid adaptive curriculum from Margolis et al. (2022)\cite{margolis2022rapid}, which increases the difficulty by adding neighboring regions to the command sampling distribution.
  \item An adaptive curriculum from Miki et al. (2022)\cite{miki2022learning}, which adjusts terrain difficulty using an adaptive method and changes elements such as reward or applied disturbances using a logistic function.
\end{itemize}

CHRL consists of curriculum learning and HER, but their contribution to controller performance still needs to be clarified. We also removed some of these components to compare their performance.

Fig. \ref{fig-Comparison}(a) shows the learning curves. In contrast to a policy without a curriculum, the learning efficiency and performance of the policy can be significantly improved regardless of the kind of curriculum learning strategy. CHRL consistently performs better and learns significantly faster than other baselines. 
The measured torques from each joint, while the robot ran at the average speed nearest 3 m/s, are shown in Fig. \ref{fig-Comparison}(b-d). CHRL has a smoother torque variation, which contributes to its high reward. 

In Fig. \ref{fig-ablation}(a), we find that the velocity tracking error of the policy increases dramatically when the velocity command is greater than 3 m/s without CHRL. This suggests that the curriculum is a crucial approach for learning high-speed locomotions. 
Other benchmarks can track high-speed motion commands stably, but have larger tracking errors than CHRL when tracking low-speed commands from 1.0 m/s to 2.0 m/s.

Fig. \ref{fig-ablation}(b) presents the CoTs versus average velocities for baselines. 
The dimensionless cost of transport (CoT)\cite{lee2020learning} is calculated to compare the efficiency of the controllers. The mechanical CoT is define as $\sum_{12 \text { actuators }}[\boldsymbol{\tau} \dot{\mathbf{q}}]^{+} /(\mathbf{m} g \mathbf{v})$, where $\mathbf{m} g$ is the total weight. 
CHRL recorded slightly lower CoTs than controllers trained with baselines. 
The presented controller is more energy efficient than the RL controller for ANYmal\cite{lee2020learning} with a log mechanical CoT of about -0.4. 



\section{Conclusions}
\label{sec:Conclusions}
We propose a framework for end-to-end training controller: Curricular Hindsight Reinforcement Learning (CHRL). We fully trained the neural network controller in simulation end-to-end with this framework. Since our controller uses only the most basic sensing, we can implement it on a low-cost robot. It is also relatively easy for others to test and improve our methods. Experimental and simulation results outline the main contributions of CHRL in learning various adaptive behaviors from experts, adapting to changing environments, and robustness to uncertainty. The experimental results show that CHRL achieves multi-modal locomotion with agility and fast response to different situations and perturbations, smooth transitions between standing balance, trotting, turning, and recovery from a fall. CHRL also enables the ability to achieve high-performance omnidirectional locomotion at high speeds. As a learning-based approach, CHRL uses computational intelligence and shows advantages in generating adaptive behaviors over traditional approaches that rely purely on explicit manual programming. However, while increasing the complexity of the task, physical simulation training may introduce some limitations. As the task becomes more complex, differences between the simulation and the real world may accumulate and become problematic. Based on the results of CHRL, future work will investigate learning algorithms that can safely refine motor skills on real hardware for more complex multi-modal tasks.





\ifCLASSOPTIONcaptionsoff
  \newpage
\fi



%

\bibliography{bare_jrnl}
\bibliographystyle{IEEEtran}

\end{document}